\documentclass{article}
\PassOptionsToPackage{numbers}{natbib}

\usepackage{booktabs}
\usepackage{multirow}
\usepackage{graphicx}
\usepackage{enumitem}
\usepackage{booktabs}
\usepackage{multirow}
\usepackage{graphicx}
\usepackage[preprint]{neurips_2026}

\usepackage[utf8]{inputenc} % allow utf-8 input
\usepackage[T1]{fontenc}    % use 8-bit T1 fonts
\usepackage{hyperref}       % hyperlinks
\usepackage{url}            % simple URL typesetting
\usepackage{booktabs}       % professional-quality tables
\usepackage{amsfonts}       % blackboard math symbols
\usepackage{nicefrac}       % compact symbols for 1/2, etc.
\usepackage{microtype}      % microtypography
\usepackage{xcolor}         % colors

\title{DEDCA: Test-Time Adaptation for Generalized AI-Generated Image Detection}

\author{%
  \textbf{Hao Tan}$^{1,2}$ \quad
  \textbf{Qimin Zhang}$^{1}$ \quad
  \textbf{Lihuang Fang}$^{3}$ \quad
  \textbf{Kebing Jin}$^{4}$ \quad
  \textbf{Jinghui Qin}$^{1}$\thanks{Corresponding author.} \\
  $^{1}$Guangdong University of Technology \\
  $^{2}$The Hong Kong University of Science and Technology (Guangzhou) \\
  $^{3}$Southern University of Science and Technology \\
  $^{4}$State Key Laboratory of Public Big Data, Guizhou University \\
  \texttt{tanhao4869@gmail.com},
  \texttt{qi-min.zhang@connect.polyu.hk}, \\
  \texttt{12432568@mail.sustech.edu.cn},
  \texttt{kbjin@gzu.edu.cn}, \\
  \texttt{scape1989@gmail.com}
}

\begin{document}

\maketitle

\begin{abstract}
AI-generated image detectors face significant challenges when deployed in real-world environments, particularly when test samples are produced by unseen generators that deviate from the source training distribution. Existing detectors often rely on a single type of evidence, such as semantic representations or low-level generation traces, making them vulnerable when the corresponding cue becomes unreliable. Moreover, adapting detectors to newly emerging generators usually requires target labels or retraining, which is costly and impractical. To address these challenges, we propose Dual-Evidence Disagreement-Constrained Adaptation (DEDCA), a test-time adaptation framework for generalized AI-generated image detection. DEDCA constructs a dual-evidence detector by combining CLIP-based semantic evidence with trace evidence extracted from shuffled entropy and residual statistics, and introduces a reliability-aware gate to perform sample-specific adaptive fusion. Our key idea is to exploit the disagreement between the two evidence streams as an unlabeled adaptation signal, enabling the detector to adjust to target-domain shifts without relying on target labels or blindly trusting a single prediction stream. Experiments show that DEDCA achieves superior performance on the GenImage benchmark by outperforming state-of-the-art AI-generated image detectors and test-time adaptation methods with 92.0\% average accuracy.
\end{abstract}

\section{Introduction}
\label{sec:intro}
\begin{figure}[t]
\centering
\includegraphics[width=0.85\linewidth]{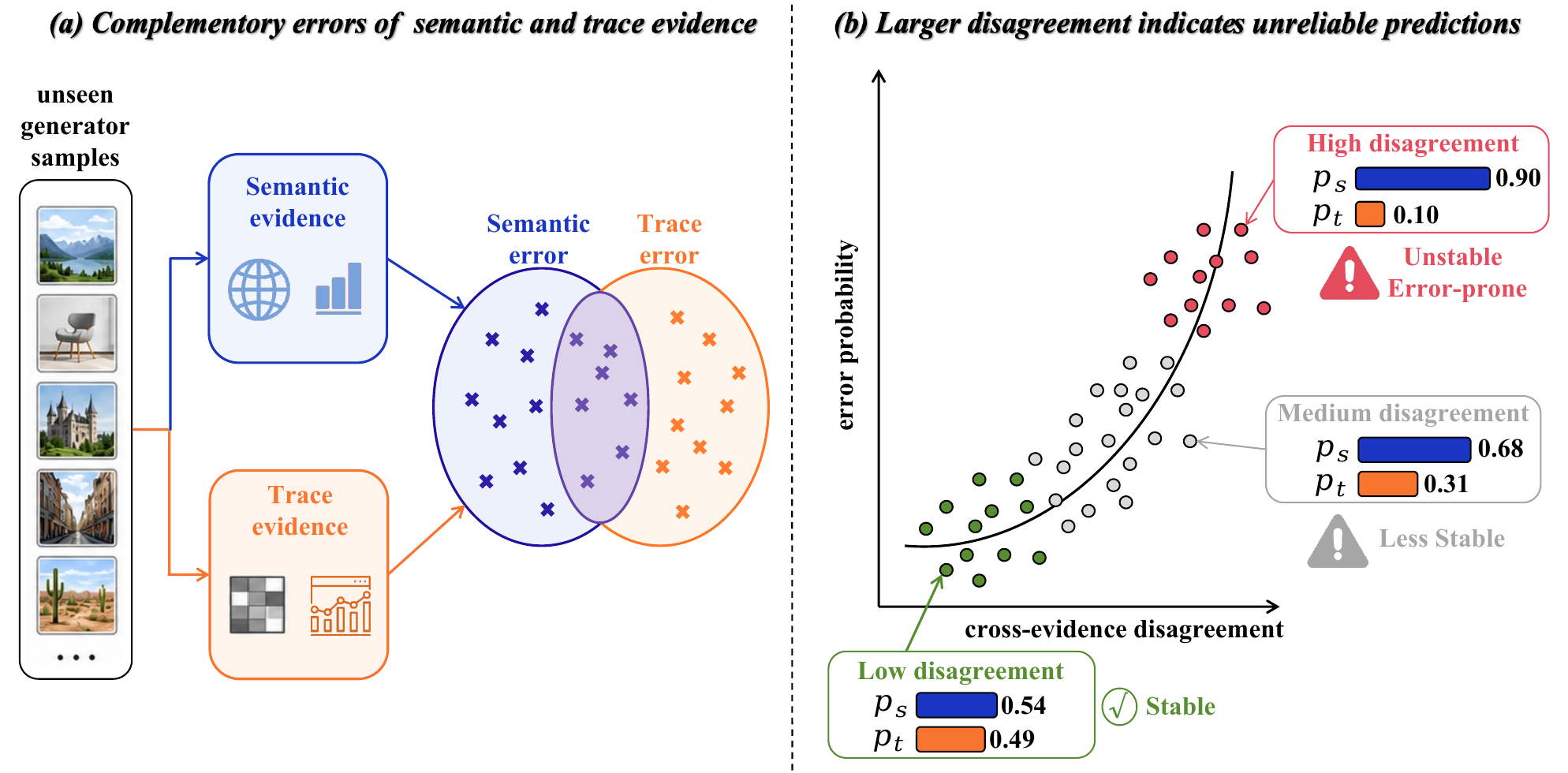}
\caption{Motivation of dual-evidence disagreement for test-time adaptation.
(a) Complementary error patterns between semantic and trace evidence. (b) Cross-evidence disagreement as an indicator of unreliable predictions.
}
\label{fig:intro}
\vspace{-6mm}
\end{figure}

Since the rapid progress of generative models~\cite{durgadevi2021generative,cao2024survey} has made synthetic images increasingly realistic and accessible, raising growing concerns about misinformation, impersonation, and the credibility of online media~\cite{chai2020makes, qin2026multimodal}, AI-generated image detection has become an important problem in visual forensics. 
Although existing detectors achieve promising results on images generated by known models~\cite{ojha2023towards, wang2023dire}, their performance often drops substantially when facing unseen generators, unseen content distributions, or unknown post-processing operations~\cite{rana2022deepfake, babaei2025generative}.
Early detectors exploit spatial or frequency artifacts produced by generative models, such as local inconsistencies, spectral traces, and high-frequency residuals~\cite{qian2020thinking, kim2025beyond}. Recent detectors improve generalization by using representations from pretrained large vision-language models~\cite{ojha2023towards, liu2024forgery}, model-based reconstruction errors~\cite{wang2023dire, ricker2024aeroblade, lin2025revisiting}, and content-suppressed low-level statistics such as neighboring pixel relationships or local entropy patterns~\cite{zhou2026beyond, yuan2025mlep}. These studies reveal two major directions for generalized image detection: 1) semantic evidence from strong pretrained visual priors, and 2) trace evidence focusing on local artifact statistics~\cite{zhang2026mff}. However, each evidence source has its own limitations. Semantic representations can transfer across content but may be affected by dataset bias and semantic shortcuts, while trace features are closer to the image formation process but can be weakened by generator-specific rendering, compression, resizing, or other post-processing operations~\cite{corvi2023detection, tan2024frequency}. This makes it difficult for a single evidence source to remain reliable under diverse deployment shifts. 
As illustrated in Fig.~\ref{fig:intro} (a), we observe that semantic-based and trace-based detectors often make different mistakes on unseen generators and their error sets are only partially overlapped, indicating that the two evidence sources capture complementary cues rather than redundant information, so a robust detector should not rely on a single evidence source, but should explicitly model both semantic-level and trace-level evidence. %Nevertheless, a naive ensemble is insufficient. Under different target shifts, the relative reliability of the two branches can change across samples. 
Furthermore, an effective detector should not only maintain two complementary evidence sources but also estimate how much each source should be trusted for a given input.

Besides, real-world deployment raises another challenge that the target distribution may continuously change~\cite{baxevanakis2022mever}. New generators, updated synthesis pipelines, and unknown post-processing operations can appear over time, but collecting newly labeled target samples to retrain is often costly and impractical~\cite{tassone2024continuous, hidayat2025use}. Test-time adaptation (TTA) offers a natural solution by adapting a pretrained model with only incoming unlabeled test samples~\cite{liang2025comprehensive}, but TTA remains much less explored in AI-generated image detection. Recent work~\cite{nguyen2025think} shows the promise of online adaptation for deepfake detection under unknown post-processing and distribution shifts, but its TTA objectives operate on a single prediction stream and rely on entropy minimization or pseudo-label-based updates~\cite{wang2020tent, niu2023towards}. However, when the source detector is unreliable on unseen target data, such self-reinforcing objectives may amplify incorrect predictions and lead to confirmation bias or model collapse. Conversely, the dual-evidence design offers a different source of supervision during adaptation. As shown in Fig.~\ref{fig:intro} (b), samples with larger disagreement between semantic and trace predictions are more likely to be misclassified or assigned unstable predictions. This observation indicates that cross-evidence disagreement can serve as an informative signal for test-time adaptation, enabling the detector to adapt to target shifts without blindly trusting a single prediction stream.

Based on this insight, we propose Dual-Evidence Disagreement-Constrained Adaptation (DEDCA), a test-time adaptation framework for generalized AI-generated image detection. Specifically, DEDCA builds a dual-evidence detector with a semantic branch and a trace branch. The semantic branch uses a frozen CLIP image encoder to extract transferable high-level visual evidence, while the trace branch applies a deterministic Shuffled Entropy-Residual Transform (SERT) to suppress semantic layout and capture generation-related local statistics from entropy maps, high-pass residuals, and gradient magnitudes. The outputs of these two branches are fused by a Decoupled Evidence Reliability Gate (DERG), which decouples their features, models bidirectional evidence interaction, and predicts sample-specific reliability weights for logit fusion. During training, we introduce a cross-evidence disagreement loss to guide dual-evidence learning, encouraging the two branches to produce compatible predictions when both evidence sources are reliable while avoiding a naive collapse into identical behavior. To adapt to a new generator, DEDCA performs Disagreement-Constrained Test-Time Adaptation on unlabeled target samples. Guided by the observation that cross-evidence disagreement indicates unreliable target predictions, DEDCA reuses this disagreement as an adaptation signal to penalize conflicts between semantic and trace predictions.
Extensive experiments on the GenImage~\cite{zhu2023genimage} benchmark show that DEDCA achieves superior performance by outperforming state-of-the-art (SOTA) AI-generated image detectors and test-time adaptation methods, consistently improving cross-generator robustness and unlabeled target-domain adaptability, especially under substantial shifts among generators.
Overall, our contributions can be summarized as follows:
\begin{itemize} %[leftmargin=*, noitemsep, topsep=0pt]
    \item We propose test-time adaptation for AI-generated image detection from a dual-evidence disagreement perspective, bringing unlabeled target-domain adaptation closer to real-world deployment.
    \item We propose DEDCA, a dual-evidence detection framework that combines semantic and trace evidence with reliability-aware fusion and exploits cross-evidence disagreement for test-time adaptation.
    \item Extensive experiments on GenImage show the generalization and adaptability of DEDCA by consistently improving robustness to unseen generators over SOTA detectors and TTA methods.
\end{itemize}

\section{Related Work}
\noindent\textbf{Semantic-based Image Detection.}
Recent AI-generated image detectors increasingly rely on high-level representations from large pretrained models. A representative line of work uses CLIP features as generalizable visual priors, where a frozen vision-language encoder with only a lightweight classifier can achieve strong cross-generator detection performance~\cite{ojha2023towards}. Following this paradigm, C2P-CLIP~\cite{tan2025c2p} further explores category-common prompts to enhance CLIP-based generalization, while recent multimodal approaches such as SIDA~\cite{huang2025sida} and AIGI-Holmes~\cite{zhou2025aigi} leverage large multimodal models for detection, localization, and explanation. Although these methods benefit from transferable semantic priors, they may still capture content bias or semantic shortcuts. In contrast, our method treats semantic representations as one evidence source and complements them with trace-level evidence.

\noindent\textbf{Trace-based Image Detection.}
Another major direction detects AI-generated images by exploiting low-level traces introduced by synthesis pipelines. Early methods focus on spatial artifacts, frequency clues, and high-frequency residuals~\cite{qian2020thinking, wang2020cnn}. Later works improve generalization through reconstruction errors from diffusion or autoencoder~\cite{wang2023dire, ricker2024aeroblade}, neighboring pixel relationships~\cite{tan2024rethinking}, and semantic-suppressed local statistics such as entropy patterns~\cite{yuan2025mlep}. More recent studies investigate bit-plane cues and bias-reduced training protocols to expose subtle generation artifacts~\cite{wang2025lota, guillaro2025bias}. These methods are closer to the generation process but remain vulnerable to post-processing perturbations. Different from them, our DEDCA treats trace evidence as a complement.

\noindent\textbf{Test-time Adaptation For Image Detection.}
Test-time adaptation updates a pretrained model using unlabeled target samples, with representative methods based on entropy minimization, augmentation consistency, sample filtering, and stable online adaptation~\cite{wang2020tent, zhang2022memo, niu2022efficient, wang2022continual, niu2023towards, zhang2024come}. Despite its relevance to real-world deployment, TTA remains less explored in AI-generated image detection. T2A~\cite{nguyen2025think} shows that online adaptation can improve deepfake detectors under unknown post-processing and distribution shifts, while benchmarks such as AIGIBench~\cite{li2025artificial} and MIRAGE~\cite{xia2025mirage} further reveal that current detectors still struggle in realistic settings. Unlike the aforementioned single-stream TTA methods, DEDCA exploits cross-evidence disagreement as an unlabeled adaptation signal, reducing dependence on pseudo-labels or entropy from one prediction stream.

\begin{figure}[t]
\centering
\includegraphics[width=0.9\linewidth]{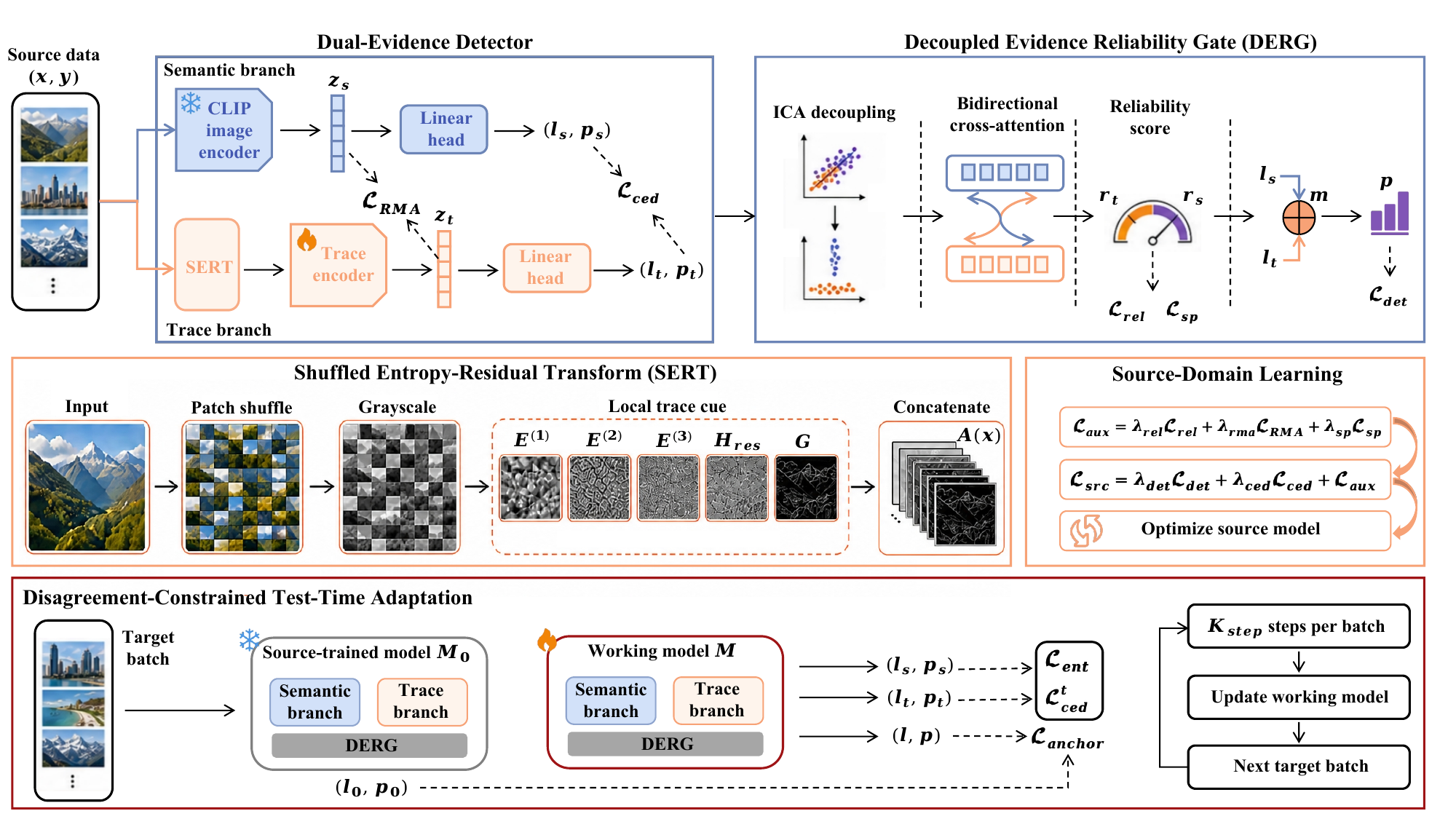}
\caption{Overall framework of DEDCA. The training stage learns a dual-evidence detector with SERT-based trace modeling and DERG-based reliability fusion, while the test-time stage adapts the detector on unlabeled target batches through disagreement-constrained optimization.
}
\label{fig:model}
\vspace{-6mm}
\end{figure}

\section{Methodology}
\label{sec:methodology}
The overall pipeline is illustrated in Fig.~\ref{fig:model}. DEDCA detects AI-generated images by jointly modeling semantic evidence and trace evidence. Given an input image, DEDCA first extracts high-level visual representations with a frozen CLIP image encoder, while a parallel trace branch applies a deterministic Shuffled Entropy-Residual Transform (SERT) to suppress global semantic layout and expose local artificial statistics. The output of these two evidence branches are then passed to a Decoupled Evidence Reliability Gate (DERG), which estimates their sample-specific reliability and fuses their logits accordingly. Besides, DEDCA uses cross-evidence disagreement in both training and test-time adaptation. During training, this disagreement regularizes semantic and trace evidence to remain compatible when both are reliable. During test-time adaptation, it serves as an unlabeled constraint for adapting the detector to target-domain shifts.

\subsection{Dual-Evidence Detector}
Given an input image $x\in \mathbb{R}^{H\times W\times 3}$, DEDCA processes it through two parallel evidence branches to produce features $(z_s, z_t)$ and logits $(\ell_s, \ell_t)$, which are subsequently used by DERG for reliability-aware fusion. 
At the semantic branch, a frozen CLIP~\cite{radford2021learning} image encoder $f_s$ and a trainable lightweight binary semantic classification head are deployed to produce required features as follows:
\begin{equation}
    z_s=f_s(x), \quad \ell_s=w_s^Tz_s+b_s, \quad p_s=\sigma(\ell_s),
\end{equation}
where $z_s\in \mathbb{R}^{d_s}$ is the semantic feature, $\ell_s$ is the semantic logit, $p_s$ is the corresponding fake probability, and $w_s$ and $b_s$ denote the weight and bias of the binary semantic classification head.

At the trace branch, the SERT is adopted to remove the original semantic layout and extract local statistical maps from the transformed image. Specifically, the image is divided into non-overlapping patches and rearranged by a fixed global permutation $\Pi$, producing a shuffled image $x_{shuf}=\Pi(x)$. The same permutation is used at all stages, including training, testing, and adaptation, so the transform is deterministic rather than an additional data augmentation. This operation can break the global object and scene layout while preserving local pixel statistics that may contain generation traces. After shuffling, SERT converts the image into a grayscale map $R\in [0,1]^{H\times W}$. Then, we quantize each grayscale intensity into one of $B$ discrete bins as $q(u,v)=min(\lfloor B\cdot R(u,v) \rfloor, B-1) \in \{0,...,B-1\}$. For each spatial location $(u, v)$, a local window $\mathcal{N}(u,v)$ is used to estimate the histogram of quantized intensities. Subsequently, for each window, the local entropy  $E(u,v)$ is defined as $E(u,v)=-\sum^{B-1}_{k=0}p_k(u,v)log(p_k(u,v)+\epsilon)$,
where $p_k(u,v)=\frac{1}{|\mathcal{N}(u,v)|}\sum_{(a,b)\in\mathcal{N}(u,v)}\mathbb{I}[q(a,b)=k]$ and $p_k(u, v)$ denote the empirical probability of the $k$-th bin in this window. This value describes the randomness of local intensity distributions. Since it depends on the distribution of quantized values rather than explicit object structures, it provides a content-suppressed cue for capturing local differences between real and generated images.
To cover trace patterns at different spatial granularities, SERT computes entropy maps at the original resolution and on two average-pooled grayscale maps. We denote the resulting maps as $E^{(1)}$, $E^{(2)}$, and $E^{(4)}$, where the superscript indicates the downsampling factor before entropy computation. In particular, $E^{(1)}$ is computed from the original grayscale map, while $E^{(2)}$ and $E^{(4)}$ are computed after $2\times$ and $4\times$ average pooling and then bilinearly upsampled back to $H\times W$. Besides, SERT also extracts two residual cues from the shuffled grayscale image: a high-pass residual map obtained by a Laplacian operator~\cite{cao2024laplace} and a gradient magnitude map obtained from horizontal and vertical Scharr responses~\cite{lin2022deep}. All these maps are normalized per sample. The final trace representation is the channel-wise concatenation of these maps as: $A(x)=Concat[E^{(1)}, E^{(2)}, E^{(4)}, H_{res}, G]$,
where $A(x) \in \mathbb{R}^{C_a\times H\times W}$, $H_{res}$ denotes the high-pass residual map, and $G$ denotes the gradient magnitude map. $A(x)$ are then fed into a convolutional trace encoder $f_t$, which is a shallow convolutional stem followed by a ResNet-18 backbone. The encoder output is passed to a linear classification head to produce the required features. This can be defined as follows:
\begin{equation}
    z_t=f_t(A(x)), \quad \ell_t=w_t^Tz_t+b_t, \quad p_t=\sigma(\ell_t),
\end{equation}
where $z_t\in \mathbb{R}^{d_t}$ is the trace feature, $\ell_t$ is the trace logit, and $w_t$ and $b_t$ are the weight and bias of the trace classification head.

\subsection{Decoupled Evidence Reliability Gate}
The semantic branch and trace branch provide complementary evidence, but their reliability is not fixed across samples. For instance, semantic evidence may be more stable when low-level traces are weakened by post-processing, whereas trace evidence may be more informative when semantic representations are affected by content bias. Therefore, DEDCA introduces a Decoupled Evidence Reliability Gate (DERG) to estimate the reliability of each branch from its feature-level interaction. Specifically, DERG first applies ICA-based~\cite{lee2020eeg} decoupling to decouple the semantic and trace features for reducing the influence of entangled feature correlations before reliability estimation as follows: $\tilde{z}_s=L2Norm((z_s-\mu_s)U_s), \quad \tilde{z}_t=L2Norm((z_t-\mu_t)U_t)$, where $U_s$ and $U_t$ are ICA unmixing matrices, and $\mu_s$ and $\mu_t$ are corresponding feature means. Then, the decoupled features $\tilde{z}_s$ and $\tilde{z}_t$ are converted into token sequences for cross-evidence interaction. Specifically, $\tilde{z}_s$ and $\tilde{z}_t$ are separately projected and reshaped into $T$ tokens with dimension $d$, denoted as $Z_s$, $Z_t\in\mathbb{R}^{T\times d}$. Subsequently,  DERG performs bidirectional cross-attention where semantic tokens query trace tokens to obtain trace-conditioned semantic context, while trace tokens query semantic tokens to obtain semantic-conditioned trace context. This procedure can be formulated as $C_s=softmax(\frac{Z_sW_q(Z_tW_k)^T}{\sqrt{d}})Z_tW_v$ and $C_t=softmax(\frac{Z_tW_q'(Z_sW_k')^T}{\sqrt{d}})Z_sW_v'$, where $C_s$ and $C_t$ are the contextualized semantic and trace tokens, respectively. $W_q$, $W_k$, $W_v$, $W'_q$, $W'_k$, and $W'_v$ are learnable projection matrices for query, key, and value transformations in the two cross-attention directions.

To estimate reliability, DERG summarizes both the original tokens and their cross-evidence contexts by average pooling. For the semantic stream, the pooled representation $[Mean(Z_s); Mean(C_s)]$ describes semantic evidence together with the trace information it attends to. Similarly, $[Mean(Z_t); Mean(C_t)]$ describes trace evidence under a semantic context. The reliability scores for two branches can be defined as $r_s=softplus(W_s^{rel}[Mean(Z_s); Mean(C_s)]+b_s^{rel})$ and $r_t=softplus(W^{rel}_t[Mean(Z_t); Mean(C_t)]+b_t^{rel})$, where $r_s$, $r_t>0$ indicate the estimated reliability scores of semantic and trace evidence, $W^{rel}_s$, $b^{rel}_s$ and $W_t^{rel}$, $b^{rel}_{t}$ are learnable parameters of the semantic and trace reliability scoring layers, respectively.  The fused logit $\ell$ is then obtained by weighting the two branch logits as follows:
\begin{equation}
    \ell = m\ell_s+(1-m)\ell_t, \quad p=\sigma(\ell),
\end{equation}
where $m=\frac{r_s}{r_s+r_t+\epsilon}$, $p$ is the fused prediction which is used as the main detection output and also serves as the basis for the training and adaptation objectives. In this way, DERG performs adaptive sample-specific evidence selection rather than static ensembling. DERG can increase the contribution of semantic evidence when trace cues are unreliable, or rely more on trace evidence when local statistical patterns provide stronger detection signals.

\subsection{Training Objective}
DEDCA is pre-trained on labeled source-domain data before test-time adaptation with two goals: 1) Each branch should learn discriminative evidence for distinguishing authentic and AI-generated images from its own feature space, and 2) The model should learn how the two evidence branches interact and coordinate due to their possibly unreliable predictions. Therefore, cross-evidence disagreement is introduced as an explicit training signal for measuring evidence conflict, allowing DEDCA to model the prediction reliability of the two branches rather than treating their predictions independently.

Given a data pair $\{(x,y)\}$, the detector produces the semantic logit $\ell_s$, trace logit $\ell_t$, and fused logit $\ell$. The basic supervised detection loss is defined over all 3 logits as follows:
\begin{equation}
    \mathcal{L}_{det}=BCE(\ell_s,y)+BCE(\ell_t,y)+BCE(\ell,y),
\end{equation}
where $BCE(\cdot,\cdot)$ denotes the binary cross-entropy loss. The branch-level terms ensure that both evidence streams remain individually discriminative, while the fused term directly optimizes the final prediction used at inference. Besides, we also introduce cross-evidence disagreement regularization to guide the interaction between the two branches via their branch-level fake probabilities $p_s$ and $p_t$, which encourages compatible predictions when both evidence sources are considered reliable, while preserving their complementary behavior when the gate detects an imbalance in reliability. It is expressed as follows:
\begin{equation}
    \mathcal{L}_{ced}=wJSD(p_{s},p_{t})^{\gamma},
\end{equation}
where $JSD(p_s,p_t)=\frac{1}{2}KL(p_s||\bar{p})+\frac{1}{2}KL(p_t||\bar{p})$,  $\bar{p}=\frac{p_s+p_t}{2}$, $\bar{p}$ denotes the mean Bernoulli prediction of the two branches, and $KL(\cdot||\cdot)$ is the Kullback--Leibler divergence. Since a direct minimization of this disagreement would force the two branches to behave similarly for all samples, which is undesirable because one evidence stream can be unreliable under certain visual conditions, the disagreement $JSD(p_s,p_t)$ is weighted by the gate confidence. The weight is defined as $w=stopgrad(4m(1-m))$, where $m$ is the semantic reliability weight predicted by DERG, and the $Stopgrad(\cdot)$ prevents the disagreement loss from trivially changing the gate weight $m$ to reduce the penalty. This weight becomes large when DERG assigns comparable reliability to both branches, and becomes small when the gate clearly favors one branch. 

Furthermore, an auxiliary calibration loss is deployed to stabilize reliability estimation and improve open-set generalization as follows:
\begin{equation}
\mathcal{L}_{aux}=\lambda_{rel}\mathcal{L}_{rel}+\lambda_{rma}\mathcal{L}_{RMA}+\lambda_{sp}\mathcal{L}_{sp},
\end{equation}
where $\mathcal{L}_{rel}=MSE(m,stopgrad(m^*))$ is the reliability supervision item, $MSE(\cdot)$ denotes the mean squared error, $m^*=\frac{exp(-e_s)}{exp(-e_s)+exp(-e_t)}$, which is the soft semantic reliability target from the semantic and trace detection errors $e_s=BCE(\ell_s,y)$ and $e_t=BCE(\ell_t,y)$ of each sample. $m^*$ is detached from the computation graph so that it can serve as a fixed sample-wise supervision signal for the gate, encouraging the gate to assign a higher weight to the branch that better explains the current labeled sample. 
$\mathcal{L}_{RMA}=\mathbb{I}(y=0)||u-u_r||^2_2+\mathbb{I}(y=1)(max\{0,\delta-||u-u_r||_2\})^2$ is the Reliability-Guided Real-Manifold Anchoring we introduced to reduce the dependence on a closed set of source fake generators. It provides an additional realness reference for the fused representation, which is useful when target fake images deviate from the source fake distribution. Real samples are encouraged to stay close to an EMA-based real image prototype $\mu_r$, while fake samples are pushed outside a margin $\delta$. $u$ is defined as $u=L2Norm(mu_s+(1-m)u_t)$, where $u_s=L2Norm(g_s(z_s))$, $u_t=L2Norm(g_t(z_t))$, $g_s$ and $g_t$ are two learnable projection heads for semantic and trace features. $\mathcal{L}_{sp}=||W_s^{rel}||_1+||W_t^{rel}||_1$ is a lightweight sparsity regularization to the reliability scoring layers, where $||\cdot||_1$ denotes the element-wise L1 norm.
Overall, the full training loss is defined as follows:
\begin{equation}    \mathcal{L}_{src}=\lambda_{det}\mathcal{L}_{det}+\lambda_{ced}\mathcal{L}_{ced}+\mathcal{L}_{aux},
\end{equation}
where $\lambda_{*}$ represents the loss weight hyperparameter.

\subsection{Disagreement-Constrained Test-Time Adaptation}
DEDCA deploys a disagreement-constrained test-time adaptation to handle the new images or domains from unseen generators or undergo unknown post-processing. Let $M_0$ denote the well-trained source model and $M$ denote the working model initialized from $M_0$. During test-time adaptation, $M_0$ is kept fixed as a source anchor, while $M$ is updated sequentially on incoming target images. To avoid unstable updates, DEDCA only updates DERG, the biases of the semantic and trace classification heads, and the affine parameters of BatchNorm layers in the trace branch. This design makes the detector adjust evidence fusion and feature normalization flexibly while preserving the learned representations at the training stage.

Specifically, given an unlabeled input $x$, the working model produces branch-level probabilities $p_{s}$, $p_{t}$ and $p_{s}$, the reliability weight $m$, and the fused prediction $p$. We first use entropy minimization on the fused prediction:
\begin{equation}
    \mathcal{L}_{ent}=-p log(p+\epsilon)-(1-p)log(1-p+\epsilon),
\end{equation}
where $\epsilon$ is a small constant for numerical stability. Then, DEDCA exploits cross-evidence disagreement as a direct unlabeled signal for target adaptation:
\begin{equation}
    \mathcal{L}_{ced}^t=w JSD(p_{s},p_{t})^{\gamma},
\end{equation}
where $w=stopgrad(4m(1-m))$. This design follows the observation that when two evidence streams trained from different cues produce conflicting predictions on a target sample, the conflict itself indicates unreliable evidence coordination under the current target distribution. $\mathcal{L}_{ced}^t$ penalizes conflicts between semantic and trace predictions when both evidence sources are considered reliable, thereby adapting the reliability gate and branch calibration to the target distribution.

However, unconstrained online adaptation can drift away from the source detector. Thus, we use the fixed source model $M_0$ to provide stable anchors and deploy an anchoring loss to constrain the shift from the source detector as follows:
\begin{equation}    \mathcal{L}_{anchor}=\lambda_{gate}\mathcal{L}_{gate}+\lambda_{drift}\mathcal{L}_{drift},
\end{equation}
where $\mathcal{L}_{gate}=||\theta_g-\theta_g^0||^2_2$ and $\mathcal{L}_{drift}=a(\ell-\ell^0)^2$. $\theta_g$ and $\theta_g^0$ denote the DERG parameters of the working model and the source model. $\mathcal{L}_{gate}$ aims to keep the DERG parameters close to their source-trained values. $\mathcal{L}_{drift}$ aims to constrain the fused logit with a confidence-aware source anchor. For each target sample, $M_0$ produces a source logit $\ell^0$ and probability $p^0$. The anchor weight $a$ can be defined as $a=stopgrad[(2|p^0-0.5|)^\beta]$, where $\beta$ controls how strongly source confidence affects the constraint. For the sample on which the source model is more confident, the $a$ will be assigned a larger value.
Overall, the final test-time adaptation objective can be defined as follows:
\begin{equation}    \mathcal{L}_{tta}=\lambda_{ent}\mathcal{L}_{ent}+\lambda_{ced}^t\mathcal{L}^t_{ced}+\mathcal{L}_{anchor},
\end{equation}
For each incoming target batch, DEDCA first computes the source predictions using $M_0$, then updates the working model $M$ for $K_{step}$ gradient steps by minimizing $\mathcal{L}_{tta}$. The updated working model produces the final predictions for the same batch and is carried forward to the next batch.

\section{Experiments}
\subsection{Experimental Setup}
\label{sec:setup}
\textbf{Datasets and  Metrics}. Following recent studies on generalized AI-generated image detection, we evaluate our method on the GenImage~\cite{zhu2023genimage} benchmark under the leave-one-generator setting. GenImage is a large-scale dataset containing real images from ImageNet and synthetic images produced by representative generative models, including Midjourney~\cite{zhu2023genimage}, Stable Diffusion V1.4~\cite{rombach2022high}, Stable Diffusion V1.5~\cite{rombach2022high}, Wukong~\cite{zhu2023genimage}, ADM~\cite{dhariwal2021diffusion}, GLIDE~\cite{nichol2021glide}, VQDM~\cite{gu2022vector}, and BigGAN~\cite{brock2018large}. In each run, one generator is held out as the unseen target domain, while the remaining generators are used as source domains for training. This protocol evaluates whether a detector can benefit from diverse source data without being confused by cross-generator interference, and whether it can still generalize to a new generator that is absent during source-domain learning. During test-time adaptation, the held-out target images are provided only as unlabeled test batches. Target labels are never used for adaptation. The working model $M$ is continuously updated along the target stream and is not reset after each batch. All compared TTA methods are evaluated under a batch-transductive setting with a batch size of 64 and 10 adaptation steps per batch. This setting reflects a realistic deployment scenario where a detector trained on known generators encounters images from a newly emerging generator without target annotations. We report accuracy (ACC) and average precision (AP) as the evaluation metrics.

\textbf{Implementation details}. We implement our method in PyTorch and conduct all experiments on a single \textit{NVIDIA RTX 3090} GPU with 24GB memory. Input images are resized to $256\times256$, while the CLIP branch follows the standard ViT-B/16 preprocessing with $224\times224$ inputs. The model is first warmed up for $K_{warm}=5$ epochs, where DERG is disabled and the semantic and trace branches are optimized with their own binary detection head, after which the FastICA transformations are fitted on source-domain features. Joint source-domain training is then performed for 50 epochs. We set $\lambda_{det}$, $\lambda_{ced}$, $\lambda_{rel}$, $\lambda_{RMA}$, and$\lambda_{sp}$ to 0.5, 10, 10, 10, and 0.1, respectively. During test-time adaptation, we update only the DERG parameters, the biases of the semantic and trace classification heads, and the affine parameters of BatchNorm layers in the trace branch. The CLIP encoder, trace convolutional weights, projection heads, and ICA transformations are kept frozen, while BatchNorm layers use fixed source-domain running statistics throughout adaptation. We deploy Adam~\cite{kingma2014adam} with a learning rate of $1e^{-3}$ and perform $K_{step}=10$ gradient steps for each incoming target batch. We set $\lambda_{ent}$, $\lambda_{ced}^t$, $\lambda_{gate}$, and $\lambda_{drift}$ to 0.5, 10, 0.2, and 0.2, respectively. We set $\gamma=2$, $\delta=1$, $\beta=2$, and $B=32$. SERT divides each $256\times256$ image into non-overlapping $8\times8$-pixel patches and uses one fixed permutation generated with seed 42 across training, validation, testing, TTA, and all leave-one-generator folds. Local entropy is computed using a $4\times4$ window with stride 1 at downsampling factors of 1, 2, and 4. Entropy, Laplacian, and Scharr responses are computed after reassembling the shuffled patches, retaining internal seams and applying reflect padding only at the outer image boundary. The semantic and trace features are both 512-dimensional and are projected into $T=16$ tokens of dimension $d=32$.

\textbf{Baselines}.
We compare the DEDCA with the multiple state-of-the-art AI-generated image detection methods and TTA methods, including \textbf{FSD}~\cite{wu2025few}, 
\textbf{AIDE}~\cite{yan2025sanity}
\textbf{MLEP}~\cite{yuan2025mlep}, \textbf{LARE2}~\cite{luo2024lare}, \textbf{MoLD}~\cite{park2025rethinking}, \textbf{PeTTA}~\cite{hoang2024persistent}, \textbf{T2A}~\cite{nguyen2025think}, and \textbf{ReservoirTTA}~\cite{vray2025reservoirtta}.

\begin{table*}[t]
\centering
\edef\DEDCApairBase{\number\value{table}}

\begin{minipage}[t]{0.68\textwidth}
\vspace{0pt}
\setcounter{table}{\numexpr\DEDCApairBase\relax}
\caption{Leave-one-generator comparison on GenImage in terms of ACC (\%). The best result is highlighted in \textbf{bold}, and the second-best result is \underline{underlined}.}
\label{tab:main_acc}
\begingroup
\scriptsize
\setlength{\tabcolsep}{2.1pt}
\renewcommand{\arraystretch}{0.92}
\resizebox{\linewidth}{!}{%
\begin{tabular}{@{}lccccccccc@{}}
\toprule
\multirow{2}{*}{\textbf{Method}} 
& \multicolumn{8}{c}{\textbf{Test Subset}} 
& \multirow{2}{*}{\textbf{\shortstack{Avg.\\Acc}}} \\
\cmidrule(lr){2-9}
& \textbf{SD1.4} & \textbf{ADM} & \textbf{BigGAN} & \textbf{GLIDE} 
& \textbf{Midj.} & \textbf{SD1.5} & \textbf{VQDM} & \textbf{Wukong} & \\
\midrule
FSD (zero-shot)      & 88.0 & 74.1 & 62.1 & 93.9 & 75.1 & 88.0 & 69.1 & 88.0 & 79.8 \\
FSD (10-shot)        & 88.8 & 79.2 & 82.2 & 97.1 & 80.9 & 88.8 & 76.2 & 88.8 & 85.2 \\
MLEP                 & 88.6 & 74.5 & 84.8 & 95.4 & 76.4 & 93.0 & 75.4 & 91.3 & 84.9 \\
LARE$^2$             & 96.2 & \textbf{82.5} & 60.1 & 96.9 & 72.4 & 90.2 & \underline{83.5} & 90.9 & 84.1 \\
AIDE             & 95.4	 & 72.8	 & 89.7 & 95.1 & \underline{82.6}	& 94.8 & 71.9	& 92.1	& 86.8\\
MoLD                 & \underline{98.5} & 77.1 & \underline{96.8} & \underline{98.7} & 81.5 & 97.4 & \textbf{84.9} & 96.0 & \underline{91.4} \\
\midrule
PeTTA                & 91.3 & \underline{81.4} & 84.2 & 91.0 & \textbf{90.2} & 92.3 & 70.3 & 90.3 & 86.4 \\
PeTTA-DEDCA          & 93.9 & 71.9 & 85.5 & 94.8 & 74.9 & 94.8 & 71.4 & 89.1 & 84.5 \\
T2A                  & 60.1 & 58.5 & 50.2 & 96.8 & 64.2 & 64.9 & 53.7 & 63.5 & 64.0 \\
T2A-MLEP             & 85.9 & 59.7 & 79.0 & 81.1 & 68.3 & 82.6 & 59.3 & 79.9 & 74.5 \\
T2A-DEDCA            & 95.9 & 76.3 & 96.1 & 96.6 & 76.1 & 96.7 & 69.1 & \underline{96.7} & 87.9 \\
ReservoirTTA-MLEP    & 78.1 & 71.3 & 61.2 & 79.8 & 73.4 & 81.1 & 57.9 & 75.9 & 72.3 \\
ReservoirTTA-DEDCA   & 84.4 & 73.9 & 86.9 & 86.2 & 73.0 & 84.9 & 68.5 & 92.4 & 81.3 \\
\midrule
DEDCA (w/o TTA)      & 97.9 & 75.7 & 95.1 & 97.7 & 80.2 & \underline{98.2} & 73.9 & 96.2 & 89.4 \\
\textbf{DEDCA (Ours)}& \textbf{99.3} & 80.2 & \textbf{98.5} & \textbf{99.1} & 82.2 & \textbf{99.1} & 78.7 & \textbf{98.7} & \textbf{92.0} \\
\bottomrule
\end{tabular}%
}
\endgroup
\end{minipage}
\hfill
\begin{minipage}[t]{0.30\textwidth}
\vspace{0pt}
\setcounter{table}{\numexpr\DEDCApairBase+2\relax}
\caption{Ablation study of the source-domain detector under the leave-ADM setting. All variants are evaluated without test-time adaptation.}
\label{tab:ablation_source}
\begingroup
\scriptsize
\setlength{\tabcolsep}{2.2pt}
\renewcommand{\arraystretch}{0.92}
\resizebox{\linewidth}{!}{%
\begin{tabular}{@{}lcc@{}}
\toprule
Setting & ACC & AP \\
\midrule
\multicolumn{3}{@{}l}{\emph{Evidence source}} \\
Semantic only & 66.8$_{\downarrow 11.8}$ & 78.6$_{\downarrow 9.1}$ \\
Trace only & 69.9$_{\downarrow 7.7}$ & 82.2$_{\downarrow 5.0}$ \\
Fixed avg. fusion & 73.0$_{\downarrow 3.6}$ & 84.9$_{\downarrow 1.8}$ \\
\midrule
\multicolumn{3}{@{}l}{\emph{SERT design}} \\
w/o patch shuffle & 74.4$_{\downarrow 1.7}$ & 85.1$_{\downarrow 1.6}$ \\
w/o multi-scale entropy & 71.7$_{\downarrow 5.3}$ & 85.4$_{\downarrow 1.3}$ \\
\midrule
\multicolumn{3}{@{}l}{\emph{DERG design}} \\
w/o ICA decoupling & 74.2$_{\downarrow 2.0}$ & 84.5$_{\downarrow 2.3}$ \\
w/o bi-cross-attn. & 72.8$_{\downarrow 3.8}$ & 83.7$_{\downarrow 3.2}$ \\
\midrule
\multicolumn{3}{@{}l}{\emph{Source objectives}} \\
w/o $\mathcal{L}_{ced}$ & 70.4$_{\downarrow 7.0}$ & 79.3$_{\downarrow 8.3}$ \\
w/o $\mathcal{L}_{rel}$ & 74.8$_{\downarrow 1.2}$ & 83.8$_{\downarrow 3.1}$ \\
w/o $\mathcal{L}_{\mathrm{RMA}}$ & 73.6$_{\downarrow 2.8}$ & 81.7$_{\downarrow 5.5}$ \\
\midrule
DEDCA full & 75.7 & 86.5 \\
\bottomrule
\end{tabular}%
}
\endgroup
\end{minipage}

\vspace{0.9em}

\begin{minipage}[t]{0.68\textwidth}
\vspace{0pt}
\setcounter{table}{\numexpr\DEDCApairBase+1\relax}
\caption{Cross-generator analysis of DEDCA on GenImage in terms of ACC/AP (\%).}
\label{tab:cross_generator}
\begingroup
\tiny
\setlength{\tabcolsep}{1.15pt}
\renewcommand{\arraystretch}{0.88}
\resizebox{\linewidth}{!}{%
\begin{tabular}{@{}llccccccccc@{}}
\toprule
\multirow{2}{*}{\textbf{Excluded}} 
& \multirow{2}{*}{\textbf{Model}}
& \multicolumn{8}{c}{\textbf{Test Subset}} 
& \multirow{2}{*}{\textbf{\shortstack{Avg.\\ACC/AP}}} \\
\cmidrule(lr){3-10}
& & \textbf{SD1.4} & \textbf{ADM} & \textbf{BigGAN} & \textbf{GLIDE} 
& \textbf{Midj.} & \textbf{SD1.5} & \textbf{VQDM} & \textbf{Wukong} & \\
\midrule
\multirow{2}{*}{SD1.4}
& DEDCA (w/o TTA) & 97.9/99.8 & 99.0/99.9 & 99.3/100 & 98.5/99.9 & 88.9/98.3 & 98.0/99.8 & 98.9/99.9 & 96.4/99.6 & 97.1/99.7 \\
& DEDCA           & \textbf{99.3/100} & 99.4/100  & 99.6/100 & 99.1/100  & 92.8/99.2 & 99.4/100  & 99.4/100  & 98.6/99.9 & 98.5/99.9 \\
\midrule
\multirow{2}{*}{ADM}
& DEDCA (w/o TTA) & 97.5/99.4 & 75.7/86.5 & 97.4/99.8 & 97.1/99.9 & 95.8/99.2 & 97.4/99.9 & 97.3/99.5 & 97.5/99.8 & 94.5/98.0 \\
& DEDCA           & 99.0/99.9 & \textbf{80.2/90.5} & 99.0/100  & 98.6/100  & 97.6/99.7 & 99.2/100  & 99.0/99.9 & 99.4/100  & 96.5/98.8 \\
\midrule
\multirow{2}{*}{BigGAN}
& DEDCA (w/o TTA) & 98.9/99.9 & 99.1/99.9 & 95.1/99.4 & 98.6/99.9 & 94.6/99.3 & 98.8/99.9 & 98.6/99.9 & 97.9/99.8 & 97.7/99.8 \\
& DEDCA           & 99.5/100  & 99.4/100  & \textbf{98.5/99.9} & 99.1/100  & 95.9/99.6 & 99.4/100  & 99.4/100  & 98.7/99.9 & 98.7/99.9 \\
\midrule
\multirow{2}{*}{GLIDE}
& DEDCA (w/o TTA) & 98.4/99.9 & 98.4/99.9 & 98.6/100 & 97.7/99.8 & 95.0/99.1 & 98.4/99.9 & 98.4/99.9 & 98.1/99.8 & 97.9/99.8 \\
& DEDCA           & 99.5/100  & 99.3/99.9 & 99.4/100 & \textbf{99.1/99.9} & 96.1/99.5 & 99.4/100  & 99.4/100  & 99.1/99.9 & 98.9/99.9 \\
\midrule
\multirow{2}{*}{Midjourney}
& DEDCA (w/o TTA) & 98.0/99.7 & 98.1/99.9 & 98.5/100 & 97.7/99.8 & 80.2/94.1 & 98.1/99.7 & 93.1/98.6 & 97.8/99.6 & 95.2/98.9 \\
& DEDCA           & 99.3/99.9 & 99.4/100  & 99.4/100 & 98.9/99.9 & \textbf{82.2/96.8} & 99.2/99.9 & 98.0/99.8 & 99.0/99.8 & 96.9/99.5 \\
\midrule
\multirow{2}{*}{SD1.5}
& DEDCA (w/o TTA) & 98.1/99.8 & 99.1/99.9 & 99.3/100 & 98.7/99.1 & 95.2/99.4 & 98.2/99.8 & 99.0/99.9 & 97.3/99.8 & 98.1/99.7 \\
& DEDCA           & 99.4/100  & 99.4/100  & 99.6/100 & 99.0/99.9 & 96.4/99.6 & \textbf{99.1/99.9} & 99.6/100  & 98.7/99.9 & 98.9/99.9 \\
\midrule
\multirow{2}{*}{VQDM}
& DEDCA (w/o TTA) & 98.8/99.9 & 98.3/99.9 & 99.1/100 & 98.2/99.8 & 90.1/98.9 & 98.7/99.9 & 73.9/85.8 & 97.2/99.6 & 94.3/98.0 \\
& DEDCA           & 99.5/100  & 99.4/100  & 99.7/100 & 99.1/99.9 & 93.9/99.5 & 99.6/100  & \textbf{78.7/89.6} & 98.3/99.9 & 96.0/98.6 \\
\midrule
\multirow{2}{*}{Wukong}
& DEDCA (w/o TTA) & 98.8/99.9 & 99.1/100 & 99.1/100 & 98.7/99.9 & 96.2/99.5 & 99.0/99.9 & 99.1/99.9 & 96.2/99.5 & 98.3/99.8 \\
& DEDCA           & 99.4/99.9 & 99.5/100 & 99.6/100 & 99.2/100  & 97.1/99.7 & 99.6/100  & 99.4/100  & \textbf{98.7/99.7} & 99.1/99.9 \\
\bottomrule
\end{tabular}%
}
\endgroup
\end{minipage}
\hfill
\begin{minipage}[t]{0.30\textwidth}
\vspace{0pt}
\setcounter{table}{\numexpr\DEDCApairBase+3\relax}
\caption{Ablation study of the test-time adaptation stage under the leave-ADM setting. All variants start from the same source-trained DEDCA checkpoint and adapt on unlabeled target batches.}
\label{tab:ablation_tta}
\begingroup
\scriptsize
\setlength{\tabcolsep}{2.2pt}
\renewcommand{\arraystretch}{0.94}
\resizebox{\linewidth}{!}{%
\begin{tabular}{@{}lcc@{}}
\toprule
Setting & ACC & AP \\
\midrule
DEDCA w/o TTA & 75.7$_{\downarrow 5.6}$ & 86.5$_{\downarrow 4.4}$ \\
$\mathcal{L}_{ent}$ only & 76.1$_{\downarrow 5.1}$ & 86.9$_{\downarrow 4.0}$ \\
w/o $\mathcal{L}_{ent}$ & 79.0$_{\downarrow 1.5}$ & 89.7$_{\downarrow 0.9}$ \\
w/o $\mathcal{L}^{t}_{ced}$ & 77.1$_{\downarrow 3.9}$ & 87.2$_{\downarrow 3.6}$ \\
w/o $\mathcal{L}_{gate}$ & 79.7$_{\downarrow 0.6}$ & 89.6$_{\downarrow 1.0}$ \\
w/o $\mathcal{L}_{drift}$ & 78.9$_{\downarrow 1.6}$ & 89.0$_{\downarrow 1.7}$ \\
w/o $\mathcal{L}_{anchor}$ & 77.9$_{\downarrow 2.9}$ & 87.8$_{\downarrow 3.0}$ \\
Update all params & 78.3$_{\downarrow 2.4}$ & 89.8$_{\downarrow 0.8}$ \\
\midrule
DEDCA full & 80.2 & 90.5 \\
\bottomrule
\end{tabular}%
}
\endgroup
\end{minipage}

\vspace{-6mm}
\end{table*}

\subsection{Main Results}
Table~\ref{tab:main_acc} shows the main experimental results. DEDCA (w/o TTA) denotes our detector without test-time adaptation, while DEDCA denotes our detector with the proposed disagreement-constrained test-time adaptation. For baselines written as TTA--detector, such as T2A-DEDCA or ReservoirTTA-MLEP, we apply the corresponding TTA algorithm to the source-only version of the detector.
As shown in Table~\ref{tab:main_acc}, DEDCA achieves the best average ACC of $92.02\pm0.68\%$ over five runs (mean $\pm$ standard deviation), compared with $91.41\pm0.94\%$ for MoLD and $89.39\pm0.52\%$ for DEDCA (w/o TTA). Both improvements are statistically significant under two-sided paired $t$-tests($p=0.021$ and $p<0.001$, respectively).
This gain indicates that the proposed adaptation stage provides consistent benefits beyond the source-trained dual-evidence detector. DEDCA obtains the best results on five out of eight target generators, including SD1.4 (99.3\%), BigGAN (98.5\%), GLIDE (99.1\%), SD1.5 (99.1\%), and Wukong (98.7\%). Although LARE$^2$ and MoLD perform better on ADM and VQDM, respectively, their performance is less balanced across different generators, e.g., LARE$^2$ drops to 60.1\% on BigGAN and MoLD reaches only 77.1\% on ADM. Conversely, DEDCA maintains strong results across both diffusion- and GAN-generated subsets, indicating stronger robustness under diverse generator shifts. Besides, those generic TTA baselines, PeTTA and ReservoirTTA, do not consistently improve their base detectors, indicating that general-purpose adaptation strategies may be unstable for AI-generated image detection. T2A, which is a task-specific TTA for AI-generated image detection, reaches 87.9\% when applied to DEDCA, but it still remains 4.1\% below DEDCA on average, showing the superiority of using cross-evidence disagreement as the adaptation signal.

The effectiveness of DEDCA can be further understood from its disagreement-driven adaptation behavior. When the semantic and trace branches conflict on an unseen target sample, the conflict indicates that at least one evidence stream is not well calibrated. DEDCA uses this conflict as an explicit optimization signal. If semantic evidence remains confident but trace evidence is weak, the cross-evidence disagreement loss encourages the adaptable trace-side parameters to match the new trace statistics. Conversely, if the trace evidence is informative but the semantic response is less reliable, DEDCA does not aggressively adapt the semantic encoder, since the frozen CLIP branch is intended to preserve a stable visual prior, and instead adjusts the reliability gate to reduce the semantic weight and rely more on the trace branch. In harder cases where one evidence source cannot be effectively corrected, the gate still provides a fallback by suppressing the unreliable branch. Therefore, DEDCA can improve target-domain performance by turning cross-evidence conflicts into a label-free calibration signal for both trace-side adjustment and reliability-aware fusion, rather than blindly enforcing self-training labels.

Furthermore, we also provide a detailed cross-generator analysis in Table~\ref{tab:cross_generator}. For each excluded generator, we evaluate DEDCA (w/o TTA) and DEDCA on all generator subsets. It can be observed that DEDCA maintains high ACC/AP on most generator subsets even when different generators are excluded from training, indicating that under our dual-evidence design, heterogeneous source generators do not introduce severe mutual interference.

\subsection{Ablation Study}
We conduct ablation studies under the leave-ADM setting, where ADM is held out as the unseen target generator. This setting is challenging enough to reveal the contribution of each component, while still allowing a clear comparison between the source-domain detector and the test-time adaptation stage. All results are reported in terms of ACC and AP.
In Table~\ref{tab:ablation_source} and Table~\ref{tab:ablation_tta} and the following ablation analysis, performance drops are reported as relative percentages with respect to the corresponding full model.
Table~\ref{tab:ablation_source} shows the ablation results of the source-domain detector. Compared with full DEDCA, using only the semantic or trace branch causes clear performance drops, with ACC decreasing by 11.8\% and 7.7\%, respectively, confirming that the two evidence sources are complementary. Directly averaging the two branches is also inferior to DERG, with a 3.6\% ACC drop, indicating that adaptive reliability estimation is more effective than static fusion. Among the architectural components, removing multi-scale entropy maps and bidirectional cross-attention leads to relatively larger ACC drops of 5.3\% and 3.8\%, showing the importance of multi-granularity trace modeling and cross-evidence interaction. For the training objectives, removing $\mathcal{L}_{ced}$ results in the largest degradation, with ACC/AP decreasing by 7.0\%/8.3\%, which indicates that the disagreement constraint is crucial for aligning semantic and trace evidence while preserving their complementary roles.

Table~\ref{tab:ablation_tta} further validates the effectiveness of the disagreement-constrained test-time adaptation. DEDCA improves over DEDCA without TTA by 5.6\% ACC and 4.4\% AP, demonstrating that unlabeled target adaptation brings consistent gains. In contrast, using $\mathcal{L}{ent}$ alone only achieves limited improvement and remains 5.1\%/4.0\% lower than the full DEDCA, indicating that confidence minimization alone is insufficient. Removing $\mathcal{L}^{t}{ced}$ causes a substantial drop of 3.9\% ACC and 3.6\% AP, confirming that target-domain cross-evidence disagreement provides the main adaptation signal. Overall, all the above results show the effectiveness of our proposed DEDCA.

\textbf{Due to space limitations, more experiments about hyperparameter analysis, model robustness, computational efficiency, and visualization analysis are reported in the technical appendices. Please refer to the appendix for more experimental results and analysis that can further validate the effectiveness of our DEDCA.}

\section{Conclusion}
In this work, we propose Dual-Evidence Disagreement-Constrained Adaptation (DEDCA), a test-time adaptation framework for generalized AI-generated image detection under unseen generators. DEDCA builds a dual-evidence detector by combining CLIP-based semantic evidence with trace evidence extracted by the Shuffled Entropy-Residual Transform (SERT). To effectively coordinate the two evidence streams, we introduce the Decoupled Evidence Reliability Gate (DERG), which estimates sample-specific reliability and performs adaptive logit fusion. Beyond source-domain learning, DEDCA further performs Disagreement-Constrained Test-Time Adaptation, where cross-evidence disagreement is used as an unlabeled adaptation signal to adjust the detector on incoming target batches without target labels. Experiments on GenImage demonstrate that DEDCA achieves stronger cross-generator generalization than existing AI-generated image detectors and TTA baselines, showing the effectiveness of exploiting complementary semantic and trace evidence and disagreement-based adaptation for generalized detection.

\section{Acknowledgments}
This work was supported in part by the National Natural Science Foundation of China (NSFC) under Grant No. 62206314 and No. 62466009, Guizhou Provincial Basic Research Program (Natural Science) under Grant No. ZK[2025]Mianshang627, the Foundation of Guizhou University under Grant No. X2024076, Science and Technology Projects in Guangzhou under Grant No. 2024A04J4388.

{
\small
\bibliographystyle{plain}
\bibliography{ref}
}

\appendix

\section{A Brief Introduction to Baselines}
We compare the DEDCA with the multiple state-of-the-art AI-generated image detection methods and TTA methods, including \textbf{FSD}~\cite{wu2025few}, 
\textbf{AIDE}~\cite{yan2025sanity},
\textbf{MLEP}~\cite{yuan2025mlep}, \textbf{LARE2}~\cite{luo2024lare}, \textbf{MoLD}~\cite{park2025rethinking}, \textbf{PeTTA}~\cite{hoang2024persistent}, \textbf{T2A}~\cite{nguyen2025think}, and \textbf{ReservoirTTA}~\cite{vray2025reservoirtta}. The brief introduction of each baseline is listed as follows:  
\begin{itemize}
    \item \textbf{FSD}~\cite{wu2025few}: It formulates generalized AI-generated image detection as a few-shot classification problem and learns a prototypical metric space, where images from the same generator are encouraged to stay close to each other.
    \item \textbf{AIDE}~\cite{yan2025sanity}: It combines CLIP-based semantic features with low-level artifact features extracted from high- and low-frequency patches selected by DCT scoring, and concatenates these features for image detection.
    \item \textbf{MLEP}~\cite{yuan2025mlep}: It constructs multi-granularity local entropy patterns from shuffled image patches at multiple scales, aiming to capture source-invariant local statistical cues while disrupting semantic content.
    \item \textbf{LARE2}~\cite{luo2024lare}: It detects diffusion-generated images by extracting latent reconstruction error features and refining image features with reconstruction-error guidance.
    \item \textbf{MoLD}~\cite{park2025rethinking}: It rethinks CLIP-ViT features for AI-generated image detection and dynamically integrates features from multiple ViT layers with a gating mechanism, rather than relying only on the final-layer representation.
    \item \textbf{PeTTA}~\cite{hoang2024persistent}: It targets persistent test-time adaptation in recurring testing scenarios and adjusts the adaptation strategy when the model tends to diverge, balancing target adaptation with collapse prevention.
    \item \textbf{T2A}~\cite{nguyen2025think}: It is an online TTA method for deepfake detection that avoids relying solely on entropy minimization by introducing uncertainty-aware negative learning, uncertain sample prioritization, and gradient masking.
    \item \textbf{ReservoirTTA}~\cite{vray2025reservoirtta}: It addresses prolonged TTA under evolving and recurring domains by maintaining a reservoir of domain-specialized models and routing incoming samples to suitable specialists through online domain clustering.
\end{itemize}

\section{Hyperparameter Analysis}

\begin{figure}[t]
\centering
\includegraphics[width=0.8\linewidth]{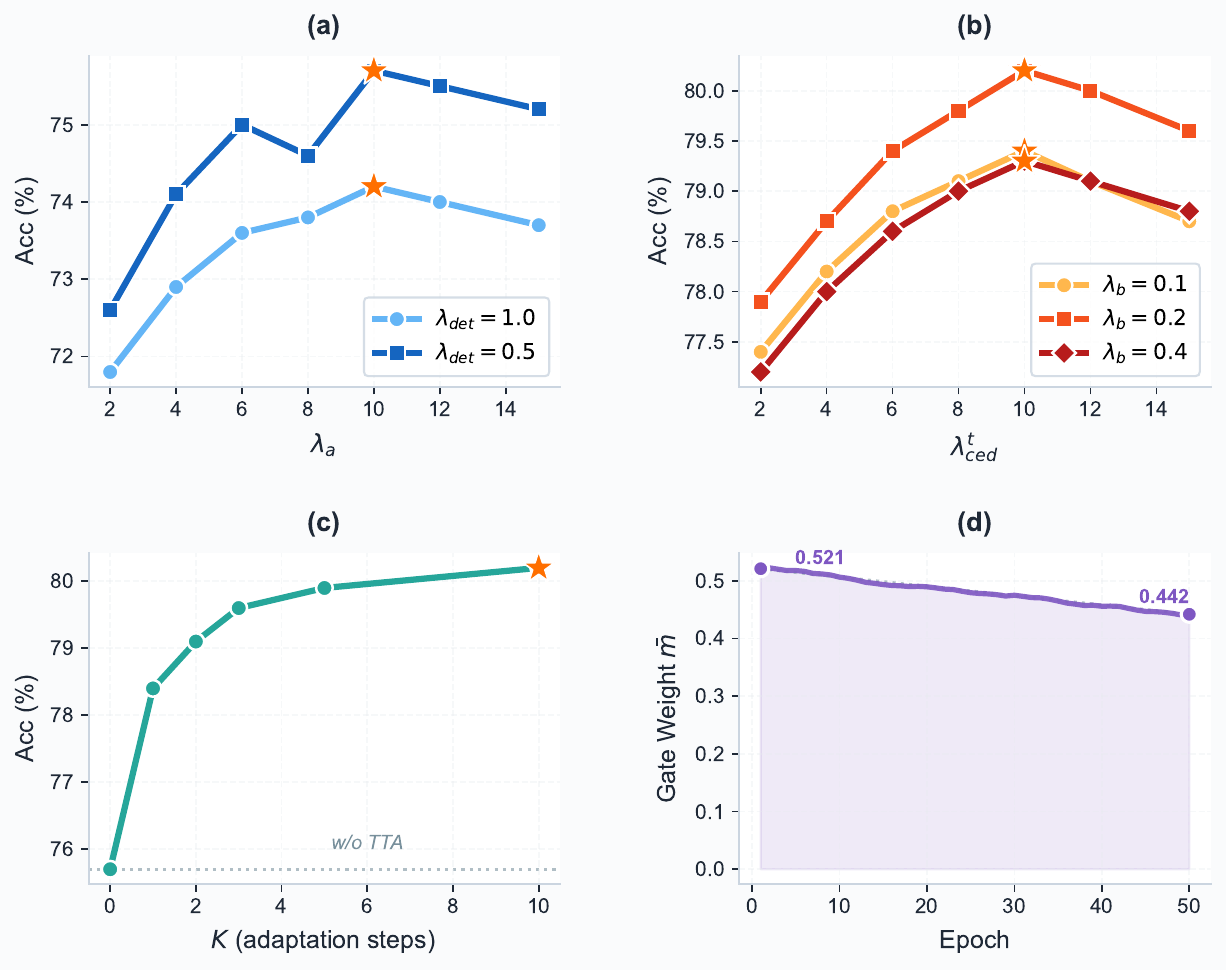}
\caption{
Parameter and gate analysis of DEDCA under the leave-ADM setting.
(a) Effect of source-domain loss balance, where $\lambda_{ced}=\lambda_{rel}=\lambda_{\mathrm{RMA}}=\lambda_a$ and $\lambda_{sp}=0.1$.
(b) Effect of test-time adaptation loss balance, where $\lambda_{ent}=0.5$ and $\lambda_{gate}=\lambda_{drift}=\lambda_b$.
(c) Effect of the number of adaptation steps $K_{step}$.
(d) Evolution of the average gate weight $\bar{m}$ during source-domain training.
}
\label{fig:parameter_gate_analysis}
\end{figure}

Figure~\ref{fig:parameter_gate_analysis} (a) studies the source-domain loss balance. We fix $\lambda_{sp}=0.1$ and set $\lambda_{ced}=\lambda_{rel}=\lambda_{\mathrm{RMA}}=\lambda_a$. It can be observed that reducing $\lambda_{det}$ from 1.0 to 0.5 consistently improves performance under different $\lambda_a$ values, suggesting that overly emphasizing the supervised detection objective may bias the model toward source-generator patterns. When $\lambda_{det}=0.5$, the ACC increases from 72.6\% to 75.7\% as $\lambda_a$ grows from 2 to 10, while larger values such as 12 and 15 bring no further gain. This indicates that stronger evidence regularization is beneficial for generalized detection, but an excessively large weight may slightly over-constrain the source-domain learning. Figure~\ref{fig:parameter_gate_analysis} (b) analyzes the TTA loss balance, where $\lambda_{ent}=0.5$ is fixed and $\lambda_{gate}=\lambda_{drift}=\lambda_b$. The best result is achieved when $\lambda_b=0.2$ and $\lambda^{t}_{ced}=10$, reaching 80.2\% ACC. In contrast, weaker anchoring with $\lambda_b=0.1$ and stronger anchoring with $\lambda_b=0.4$ both lead to lower performance, showing that test-time adaptation requires a proper balance between target-domain disagreement optimization and source-model preservation.

Figure~\ref{fig:parameter_gate_analysis} (c) shows the effects of the number of adaptation steps. Compared with the non-adapted baseline, using only one adaptation step improves ACC from 75.7\% to 78.4\%, indicating that DEDCA can quickly exploit unlabeled target data. The performance finally increases to 80.2\% at $K=10$. Figure~\ref{fig:parameter_gate_analysis} (d) visualizes the evolution of the epoch-averaged reliability weight $\bar{m}$ during source-domain training. Since $m$ denotes the fusion weight of the semantic branch and $1-m$ corresponds to the trace branch, the decrease from 0.521 to 0.442 indicates that DERG gradually shifts the fused prediction from a slight preference for semantic evidence to a stronger reliance on trace evidence. This observation provides an additional insight: semantic cues may be more useful for establishing a transferable representation at the early stage, whereas generation-related trace cues become more reliable after sufficient cross-evidence optimization.

\section{Robustness Evaluation}

To evaluate robustness under realistic image transformations, we apply common post-processing operations to the held-out test sets and report the average results over all eight leave-one-generator settings. Specifically, the post-processing operations include color saturation adjustment with factor 2.0, color contrast adjustment with factor 0.6, Gaussian blur with $\sigma=1.0$, resizing with scale factor 0.75, and JPEG compression with quality factor 75. These operations are applied to both real and fake samples.

\begin{table*}[t]
\centering
\caption{
Robustness evaluation under common post-processing operations.
$\Delta$ denotes the absolute ACC drop compared with the clean average performance of each method.
}
\label{tab:robustness_postprocessing}
\setlength{\tabcolsep}{3.6pt}
\renewcommand{\arraystretch}{1.12}
\resizebox{\textwidth}{!}{
\begin{tabular}{l|ccc|ccc|ccc|ccc|ccc|ccc}
\toprule
\multirow{2}{*}{Method}
& \multicolumn{3}{c|}{Color Saturation}
& \multicolumn{3}{c|}{Color Contrast}
& \multicolumn{3}{c|}{Gaussian Blur}
& \multicolumn{3}{c|}{Resize}
& \multicolumn{3}{c|}{JPEG}
& \multicolumn{3}{c}{Average} \\
\cmidrule(lr){2-4}
\cmidrule(lr){5-7}
\cmidrule(lr){8-10}
\cmidrule(lr){11-13}
\cmidrule(lr){14-16}
\cmidrule(lr){17-19}
& ACC & AP & $\Delta$
& ACC & AP & $\Delta$
& ACC & AP & $\Delta$
& ACC & AP & $\Delta$
& ACC & AP & $\Delta$
& ACC & AP & $\Delta$ \\
\midrule
MoLD
& 89.0 & 96.8 & 2.4
& 88.1 & 96.0 & \textbf{3.3}
& 86.7 & 94.8 & \textbf{4.7}
& 87.2 & 95.2 & \textbf{4.2}
& 86.9 & 95.5 & 4.5
& 87.6 & 95.7 & \textbf{3.8} \\

T2A-DEDCA
& 83.9 & 94.0 & 4.0
& 82.5 & 92.8 & 5.4
& 80.6 & 91.2 & 7.3
& 81.0 & 91.6 & 6.9
& 81.7 & 92.1 & 6.2
& 81.9 & 92.3 & 6.0 \\

ReservoirTTA-DEDCA
& 77.5 & 90.6 & 3.8
& 77.0 & 89.2 & 4.3
& 75.1 & 87.5 & 6.2
& 74.9 & 88.1 & 6.4
& 76.4 & 88.6 & 4.9
& 76.2 & 88.8 & 5.1 \\

DEDCA (w/o TTA)
& 85.2 & 96.1 & 4.2
& 83.4 & 94.8 & 6.0
& 81.0 & 92.6 & 8.4
& 81.8 & 93.2 & 7.6
& 82.5 & 93.9 & 6.9
& 82.8 & 94.1 & 6.6 \\

\midrule
DEDCA
& \textbf{90.4} & \textbf{98.0} & \textbf{1.6}
& \textbf{88.6} & \textbf{97.0} & 3.4
& \textbf{86.8} & \textbf{95.9} & 5.2
& \textbf{86.7} & \textbf{95.6} & 5.3
& \textbf{88.0} & \textbf{96.7} & \textbf{4.0}
& \textbf{88.1} & \textbf{96.6} & 3.9 \\
\bottomrule
\end{tabular}}
\end{table*}

Table~\ref{tab:robustness_postprocessing} reports the robustness evaluation results. DEDCA achieves the best average ACC/AP of 88.1\%/96.6\% across all post-processing operations, outperforming DEDCA (w/o TTA) by 5.3\% ACC and 2.5\% AP. It also reduces the average ACC drop from 6.6 to 3.9, showing that the proposed adaptation can effectively compensate for distribution shifts introduced by post-processing. DEDCA achieves the highest ACC of 86.8\% under Gaussian blur and remains competitive under resizing with an ACC of 86.7\%. MoLD shows competitive robustness in terms of ACC drop, which is reasonable since its CLIP-based multi-layer representation is less directly tied to low-level trace statistics. However, its absolute average ACC/AP remains lower than DEDCA. Compared with generic TTA baselines such as T2A-DEDCA and ReservoirTTA-DEDCA, DEDCA consistently achieves stronger performance, indicating that disagreement-constrained adaptation is more suitable for robust generalized AI-generated image detection.

\section{Computational Efficiency}
\label{sec:comp}
We further evaluate the computational efficiency of DEDCA to examine the additional cost introduced by the dual-evidence detector and the disagreement-constrained test-time adaptation. Table~\ref{tab:computational_cost} reports the model size, the number of parameters updated during test-time adaptation, the non-TTA inference time, and the adaptation time. All measurements are conducted on a single \textit{NVIDIA RTX 3090} GPU.  It can be observed that DEDCA achieves a favorable balance between detection performance and computational cost. Compared with MoLD, DEDCA uses fewer parameters and reduces the total test-time cost from 10.72 ms to 7.38 ms even when its adaptation time is included, while improving the average ACC from 91.4\% to 92.0\%, corresponding to a relative gain of about 0.7\%. Compared with MLEP, DEDCA introduces a larger model and higher runtime due to its dual-evidence design, but improves the average ACC from 84.9\% to 92.0\%, giving a relative gain of about 8.4\%. Compared with LARE$^2$, DEDCA has slightly fewer parameters, although its inference time is higher, and improves the average ACC from 84.1\% to 92.0\%, corresponding to a relative gain of about 9.4\%. During test-time adaptation, DEDCA updates 540.3K parameters, including only a restricted subset of the model, while keeping the main feature extractors frozen. Although this number is larger than that of the compared TTA methods, its TTA time remains moderate at 4.54 ms and is lower than that of T2A. Compared with PeTTA, DEDCA only increases the TTA time from 3.63 ms to 4.54 ms, while improving the average ACC from 86.4\% to 92.0\%, giving a relative gain of about 6.5\%. These results indicate that DEDCA's dual-evidence mechanism obtains stronger detection performance without excessively increasing the computational burden.

\begin{figure}[t]
\centering
\includegraphics[width=\linewidth]{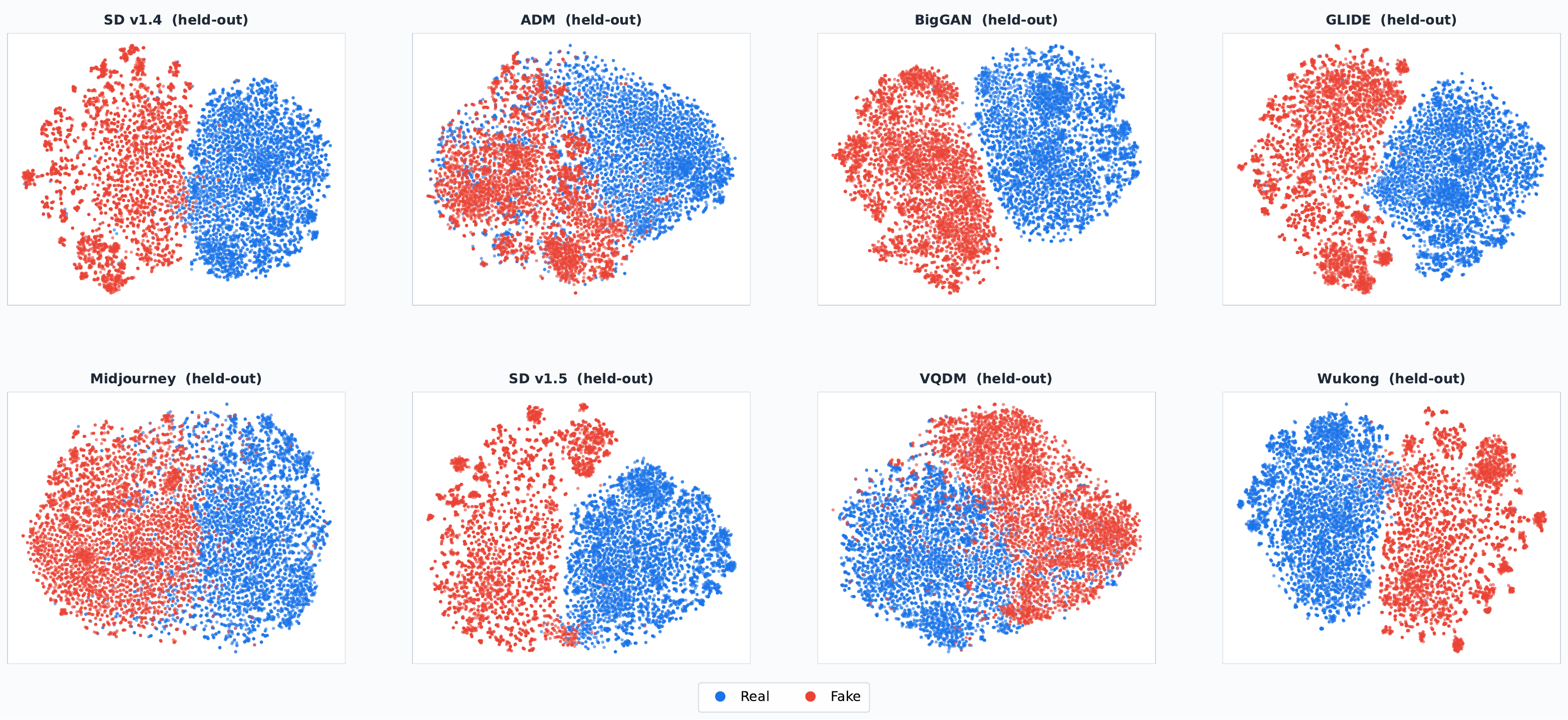}
\caption{
t-SNE visualization of real and fake feature distributions produced by DEDCA (w/o TTA).
Each subplot corresponds to one held-out generator, where blue and red points denote real and fake samples, respectively.
}
\label{fig:tsne_deca}
\end{figure}

\begin{figure}[t]
\centering
\includegraphics[width=\linewidth]{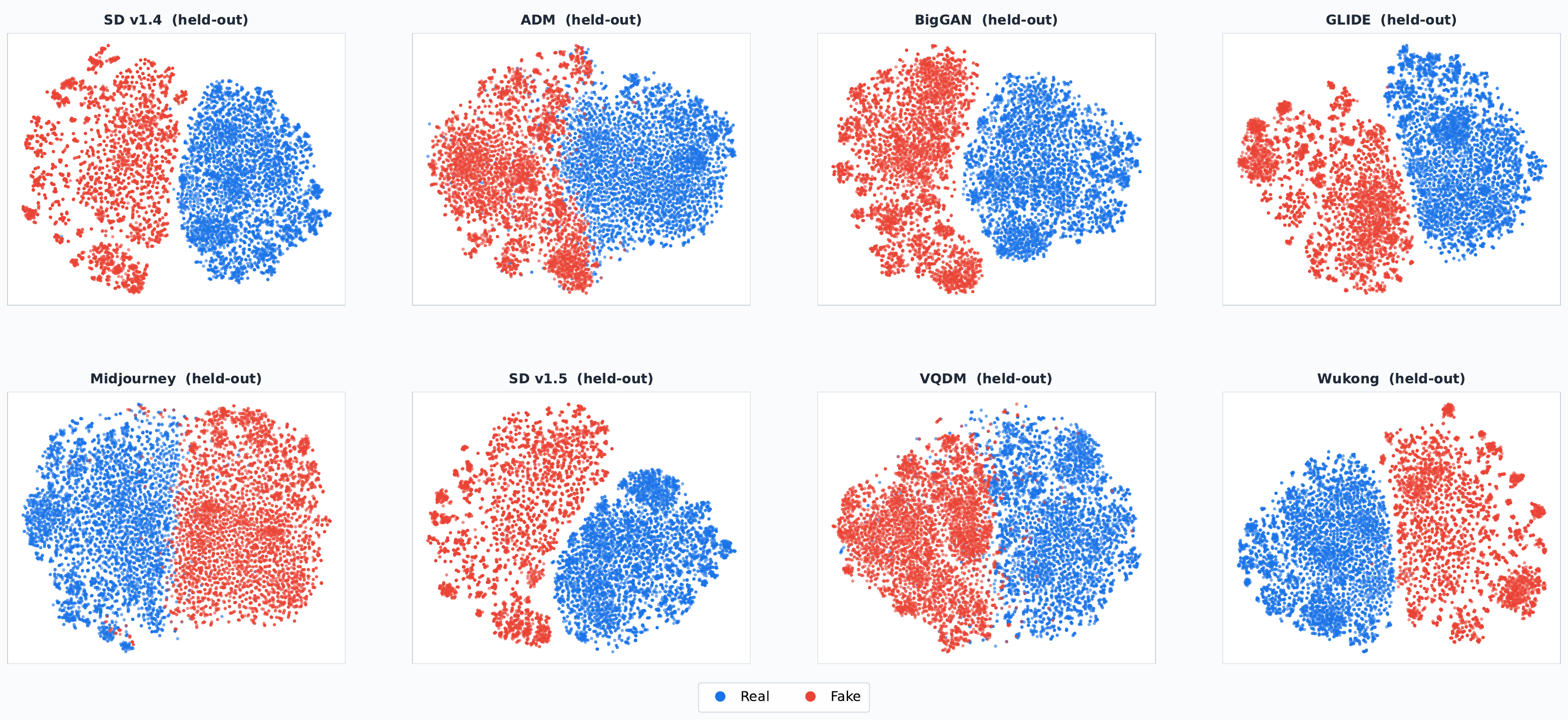}
\caption{
t-SNE visualization of real and fake feature distributions produced by DEDCA.
Each subplot corresponds to one held-out generator, where blue and red points denote real and fake samples, respectively.
}
\label{fig:tsne_dedca}
\end{figure}

\begin{figure}[t]
\centering
\includegraphics[width=\linewidth]{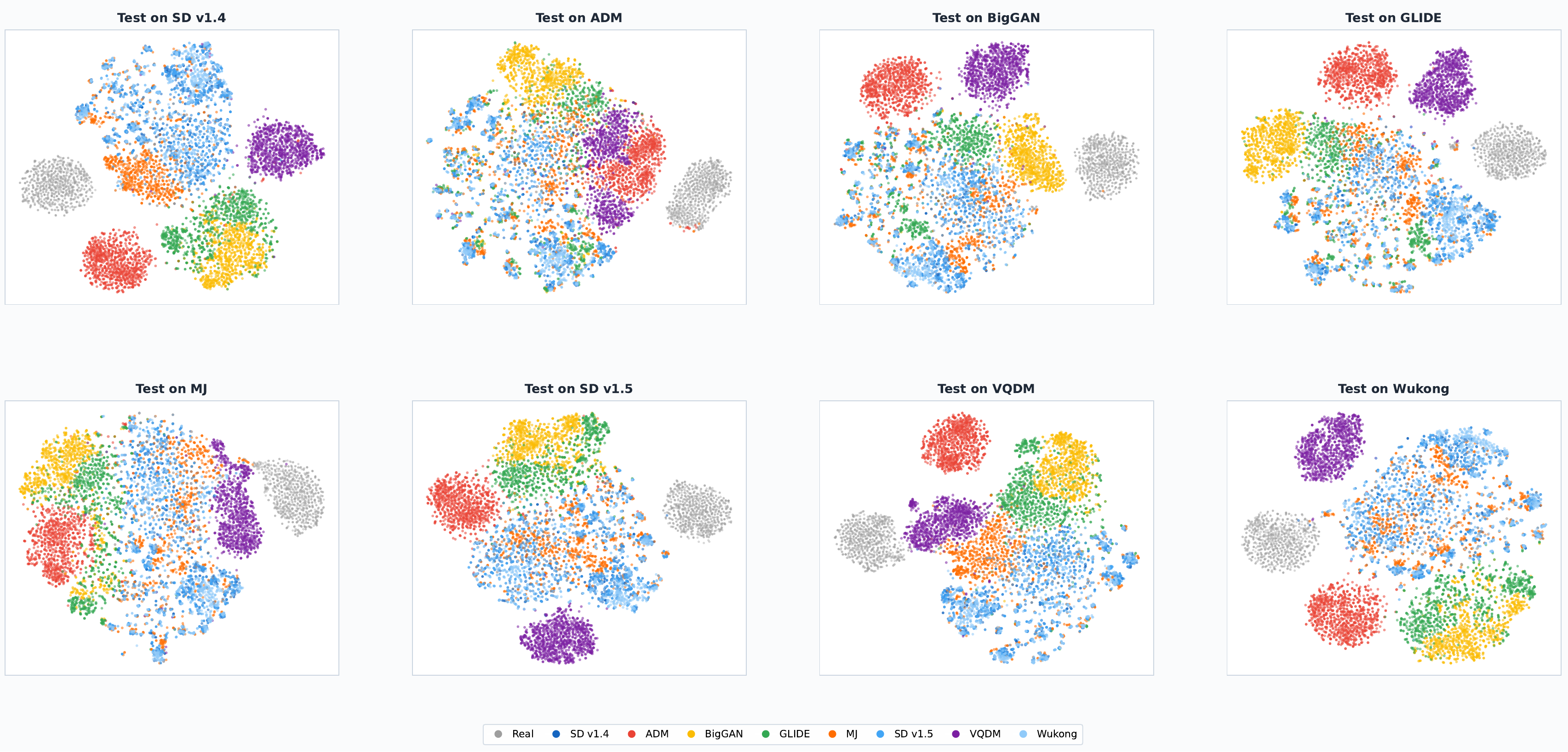}
\caption{
Multi-generator t-SNE visualization of the feature space produced by DEDCA without test-time adaptation.
}
\label{fig:tsne_multi_deca}
\end{figure}

\begin{figure}[t]
\centering
\includegraphics[width=\linewidth]{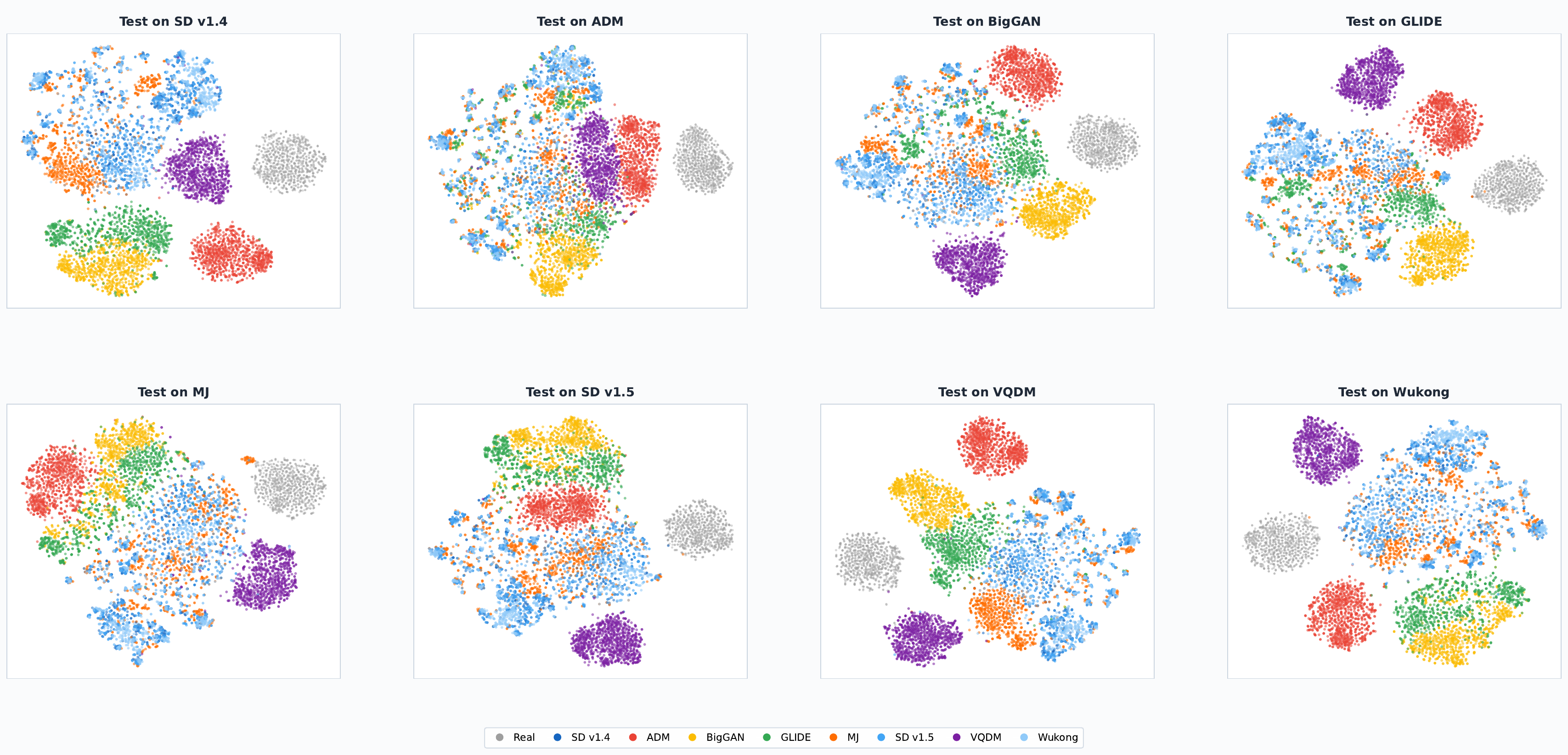}
\caption{
Multi-generator t-SNE visualization of the feature space produced by DEDCA with test-time adaptation.
}
\label{fig:tsne_multi_dedca}
\end{figure}

\begin{table}[t]
\centering
\caption{Computational cost comparison. Infer Time denotes the non-TTA inference time. Params@TTA denotes the number of parameters updated during test-time adaptation.}
\label{tab:computational_cost}
\small
\setlength{\tabcolsep}{6.0pt}
\begin{tabular}{lcccc}
\toprule
Method & Total Params & Params@TTA & Infer Time & TTA Time \\
\midrule
MLEP                         & 1.40M   & --     & 1.42  & --    \\
LARE$^2$                     & 104.31M & --     & 0.85  & --    \\
MoLD                         & 303.71M & --     & 10.72 & --    \\
Xception                     & 20.81M  & --     & 1.57  & --    \\
T2A (Xception)               & 20.81M  & 54.5K  & 1.57  & 11.70 \\
PeTTA (ResNet-50)            & 25.56M  & 53.1K  & 0.65  & 3.63  \\
ReservoirTTA (ResNet-50)     & 25.56M  & 53.1K  & 0.65  & 3.02  \\
DEDCA (w/o TTA)              & 98.07M  & --     & 2.84  & --    \\
DEDCA (Ours)                 & 98.07M  & 540.3K & 2.84  & 4.54 \\
\bottomrule
\end{tabular}
\end{table}

\section{Visualization Analysis}
To further evaluate the effectiveness of the DEDCA, we visualize the feature spaces of DEDCA (w/o TTA) and DEDCA using t-SNE. The feature distributions of real and fake samples from the held-out generator under different leave-one settings are shown in Figure~\ref{fig:tsne_deca} and Figure~\ref{fig:tsne_dedca}. It can be observed that the two categories form clear separations in the feature space, indicating that DEDCA can learn discriminative real-fake representations even when facing generators unseen during training. Compared with DEDCA (w/o TTA), the DEDCA further produces more compact intra-class distributions and clearer real-fake margins, showing that test-time adaptation improves the target-domain feature structure.

Furthermore, Figure~\ref{fig:tsne_multi_deca} and Figure~\ref{fig:tsne_multi_dedca} visualize all eight generators in the same feature space for DEDCA (w/o TTA) and DEDCA. In both cases, real samples generally form a distinguishable cluster, while fake samples exhibit meaningful generator-related structures rather than collapsing into a single mixed distribution. ADM and VQDM tend to occupy relatively independent regions, and BigGAN and GLIDE also appear as recognizable groups. SD v1.4, SD v1.5, and Wukong are naturally located in adjacent regions, since they share closely related diffusion-based generation characteristics. Midjourney is also close to this group, which may be attributed to its similar photorealistic rendering style and diffusion-like local trace patterns. After test-time adaptation, these generator-related structures become more organized while maintaining better separation from real samples. This observation suggests that DEDCA does not merely learn a binary real-fake boundary, but also implicitly preserves generator-family-level trace variations in the learned representation.

\section{Current Limitations and Future Work}
\label{sec:limit}
While DEDCA demonstrates strong cross-generator generalization and effective unlabeled test-time adaptation, several directions remain worth further exploration. First, our experiments mainly follow the leave-one-generator protocol on GenImage, which provides a controlled evaluation of unseen-generator robustness. 
Future work could further validate the method in more open-world scenarios involving mixed generators, continuously updated synthesis pipelines, user editing, screenshots, and platform-specific image transformations. 
Further evaluation could also extend the current batch-transductive protocol to randomized target-stream orders and strictly prequential predict-then-adapt settings.
Second, although SERT provides useful content-suppressed trace evidence, studying its behavior under more diverse post-processing and editing operations would offer a more comprehensive understanding of practical robustness. Third, DEDCA currently performs online adaptation with a restricted set of trainable parameters, and future work may explore richer reliability modeling, memory-assisted adaptation, and lightweight dual-evidence architectures to further improve long-term stability and deployment efficiency.

\end{document}